\documentclass[review,5p,twocolumn]{elsarticle}
\pdfoutput=1
\usepackage{lineno,hyperref}
\usepackage{amsmath}
\usepackage{amssymb}
\hypersetup{pdfauthor=author}
\modulolinenumbers[5]

\newcommand{\etal}{\textit{et al}.~}

\newcommand{\ieno}{\textit{i}.\textit{e}.}

\newcommand{\egno}{\textit{e}.\textit{g}.} 

\newcommand{\etcno}{\textit{etc}}

\journal{Journal of \LaTeX\ Templates}

\begin{document}

\begin{frontmatter}

\title{Semantically Video Coding: Instill Static-Dynamic Clues \\ into Structured Bitstream for AI Tasks}


\author[mymainaddress]{Xin Jin\corref{equalcontribution}}
\ead{jinxustc@mail.ustc.edu.cn}

\author[mymainaddress]{Ruoyu Feng\corref{equalcontribution}}
\ead{ustcfry@mail.ustc.edu.cn}

\author[mymainaddress]{Simeng Sun\corref{equalcontribution}}
\ead{smsun20@mail.ustc.edu.cn}

\author[mymainaddress]{Runsen Feng}
\ead{fengruns@mail.ustc.edu.cn}

\author[mysecondaryaddress]{Tianyu He}
\ead{ deeptimhe@gmail.com }

\author[mymainaddress]{Zhibo Chen, Corresponding Author\corref{mycorrespondingauthor}}
\ead{chenzhibo@ustc.edu.cn}


\cortext[equalcontribution]{Equal contribution.}
\cortext[mycorrespondingauthor]{Corresponding author.}

\address[mymainaddress]{University of Science and Technology of China}
\address[mysecondaryaddress]{Alibaba Group}

\begin{abstract}
Traditional media coding schemes typically encode image/video into a semantic-unknown binary stream, which fails to directly support downstream intelligent tasks at the bitstream level. Semantically Structured Image Coding (SSIC) framework~\cite{sun2020semantic} makes the first attempt to enable decoding-free or partial-decoding image intelligent task analysis via a Semantically Structured Bitstream (SSB). However, the SSIC only considers image coding and its generated SSB only contains the static object information. In this paper, we extend the idea of semantically structured coding from video coding perspective and propose an advanced Semantically Structured Video Coding (SSVC) framework to support heterogeneous intelligent applications. Video signals contain more rich dynamic motion information and exist more redundancy due to the similarity between adjacent frames. Thus, we present a reformulation of semantically structured bitstream (SSB) in SSVC which contains both of static object characteristics and dynamic motion clues. Specifically, we introduce optical flow to encode continuous motion information and reduce cross-frame redundancy via a predictive coding architecture, then the optical flow and residual information are reorganized into SSB, which enables the proposed SSVC could better adaptively support video-based downstream intelligent applications. Extensive experiments demonstrate that the proposed SSVC framework could directly support multiple intelligent tasks just depending on a partially decoded bitstream. This avoids the full bitstream decompression and thus significantly saves bitrate/bandwidth consuming for intelligent analytics. We verify this point on the tasks of image object detection, pose estimation, video action recognition, video object segmentation, etc.
\end{abstract}


\end{frontmatter}

\linenumbers

\section{Introduction}

The multimedia industry, where image/video content plays a pivotal role, is developing rapidly. The emergence of next-generation mobile networks will bring greater opportunities and challenges to the traditional multimedia industry. Meanwhile, with the human society moving from informatization to intelligence, more and more image/video intelligent applications are applied to public safety monitoring, autonomous driving, remote machine control, Internet medical treatment, military defense, \etcno. In the above-mentioned scenarios, it is necessary to ensure the interpretability and interoperability of intelligent analysis results. Therefore, introducing new multimedia analytics paradigms for machine intelligence is attracting more and more attention. This will become an important development trend of artificial intelligence in the future. 

\begin{figure}
  \centerline{\includegraphics[width=1.0\linewidth]{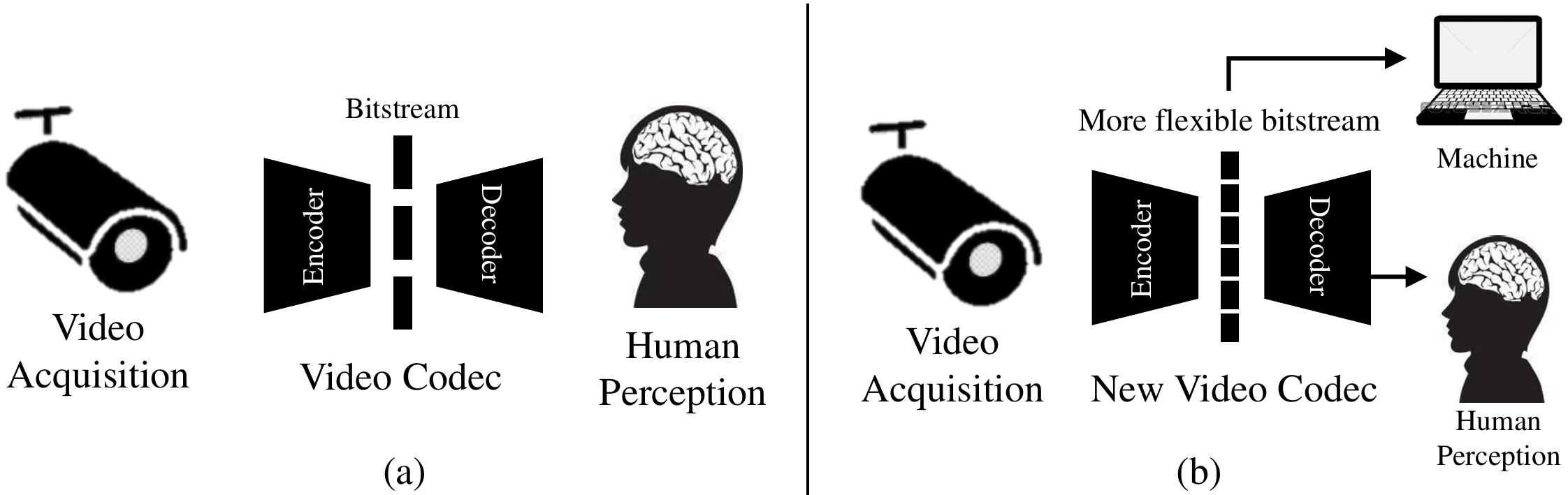}}
    \caption{Motivation illustration: (a) the traditional coding frameworks typically only focus on satisfying human perception. (b) as a high-efficient coding framework that serves for the AI era, it should satisfy both human perception and machine analytics.}
\label{fig:motivation_1}
\end{figure}

As a pivotal role in the modern multimedia industry, video occupies most of the communication bandwidth. To alleviate transmission burden and save storage resources, the video content is typically compressed into a compact representation (\ieno, bitstream), during the transmission procedure~\cite{SP2, SP3, SP4}. Once the raw video content information needs displaying for human eyes or employing for multimedia analysis applications, a reverse decoding operation will be applied to recover such compact representations to the raw pixel deployment. In specific, the traditional hybrid video coding (HVC) frameworks~\cite{forchheimer1981differential} have evolved over the decades with gradually integrating the efficient transformation, quantization, and entropy coding
into the compression procedure, achieving a trade-off on the rate-distortion optimization~\cite{standardquantization}. Particularly, the mainstream traditional video coding frameworks, \egno, MPEG-4 AVC/H.264~\cite{wiegand2003overview,wang2014efficient}, High
Efficiency Video Coding (HEVC)~\cite{sullivan2012overview,zhu2014screen,zhang2017ctu,tsang2018reduced} and the recently-proposed versatile video coding (VVC)~\cite{standardquantization}, have achieved great success. They all try to promise less distortion between the raw
image and reconstructed image with lower bit-rate cost. 

For supporting the fast-developing intelligent tasks, these traditional codecs need to fully decode all the compressed code streams to reconstruct the raw data. However, the decoding procedure of existing HVC codecs is inevitably faced with unexpected high computational complexity and large time consuming. This severely restricts the practical applications of these coding schemes. For example, in order to support the machine learning based multimedia algorithms, \egno, detection, recognition, tracking, \etcno, the traditional coding frameworks typically need first decompresses all the encoded bitstream into the raw RGB/YUV format, and then feed the decompressed video content into downstream tasks for the further analysis, which inevitably consumes a large amount of decoding computations
when meeting the large-scale intelligent media applications at the edge server side (terminal). Therefore, as shown in Fig.~\ref{fig:motivation_1}, a high-efficient coding framework should compress the captured media content into a more flexible format, which not only can be perceived by humans through data decompression, but also can be directly handled by machine learning algorithms with much less decompression complexity or even no decompression procedure. This could significantly save the bitstream transmission and decoding cost. Recently, MPEG has also initiated the standard activity on video coding
for machine (VCM)~\footnote{https://lists.aau.at/mailman/listinfo/mpeg-vcm}, which attempts to identify the opportunities and
challenges of developing collaborative compression techniques for humans and machines, while establishing a new coding standard for both machine vision and hybrid machine-human vision scenarios.

\begin{figure}
  \centerline{\includegraphics[width=1.0\linewidth]{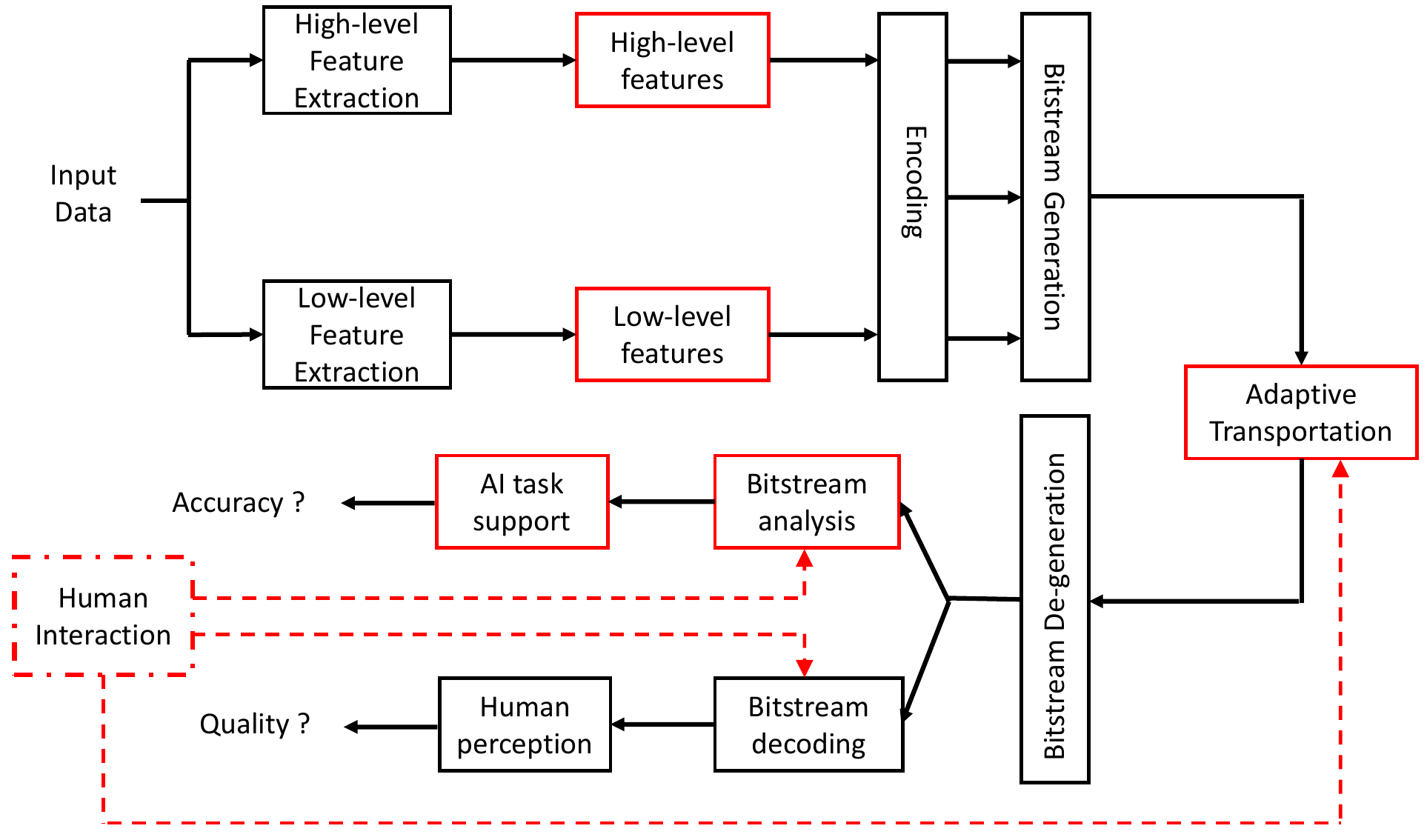}}
    \caption{Overview of our idea. The boxes marked as red denote the novel designs compared to existing codecs. In specific, these red solid boxes mean the new designs that aim to achieve semantically structured bitstream (SSB) for machine analytics.}
\label{fig:overview}
\end{figure}

In recent years, with the fast development of the deep learning based compression techniques~\cite{balle2017end,balle2018variational,chen2019learning}, several studies contribute some new compression schemes that could directly support downstream intelligent tasks without decoding all the compressed bitstream~\cite{duan2020video}. Torfason~\etal,~\cite{torfason2018towards} use a neural network to generate a compressed bitstream as input for supporting the downstream tasks directly, such as classification and segmentation, which bypasses decoding of the compressed representation into RGB space, and thus reducing computational cost. Similar ideas can be found in the video-based schemes, CoViAR~\cite{wu2018compressed} and DMC-Net~\cite{shou2019dmc}, they directly leverage the motion vectors and residuals that are both readily available in the compressed video to represent motion at no cost to support the downstream action recognition task. However, these schemes are still task-specific, or said, designed for a limited range of applications, they cannot meet the higher and more general requirements for flexibility and efficiency. Because they did not consider the intrinsic semantics contained in the compressed bitstream, and can not leverage different structural bitstream for different tasks.

Sun~\etal~\cite{sun2020semantic} first introduce a new concept of semantically structured coding for image compression field (abbreviated as SSIC), and generate a semantically structured bitstream (SSB), where each part of the bitstream represents a specific object and can be directly used for the aforementioned intelligent image tasks (including object detection, pose estimation~\etcno.) However, this work only considers the image coding framework, the generated SSB only contains the static object information of the image, which seriously limits its practical application to a larger scope, especially for video-based intelligent applications.

Therefore, in this paper, we extend the idea of semantically structured coding from video coding perspective and propose a new paradigm of video coding
for machines (VCM). Specifically, we introduce an advanced Semantically Structured Video Coding (SSVC) framework to directly support heterogeneous intelligent multimedia applications. As illustrated in Fig.~\ref{fig:overview}, in order to generate a semantic-sensing bitstream that could be directly used for supporting downstream intelligent analytics without fully decoding and also could be reconstructed for human perception, SSVC codec encodes the input media data (\ieno, image or video) into a semantically structured bitstream (SSB). SSB generally consists of hierarchical information of \emph{high-level features}, (\egno, the category and spatial location information of each object detected in the video) and \emph{low-level features}, (\egno, the content information of each object or the rest background in the video).



In detail, for these video key frames of the intra-coded frames, we herein leverage a simple and effective object detection technique to help instantiate the static information of SSB. We integrate the recently proposed CenterNet~\cite{zhou2019objects} in the encoder of our SSVC framework, which aims to locating objects and obtaining their corresponding class ID and spatial location (\egno, bbox) information in the feature domain. Then, we re-organize such features to form a part of SSB, by which some specific objects can be reconstructed and several image-based intelligent analysis tasks such as object classification/detection could achieve similar or better results than fully-decompressed images. 


Except for the static semantic information that derived from objects of i-frames, the motion characteristics is also very improtant for video compression~\cite{mao2019convolutional}. Therefore, our SSVC further integrates motion clues, denoted by optical flow and content residues, of the continuous video frames (\ieno, p-frames, inter-coded using reference frames from the past) into SSB to enable a wider of video tasks supporting. For example, for a video-based multimedia intelligent analysis task, \egno, video action recognition, only the person-related content of key frame (\ieno, i-frame) and the corresponding optical flow of continuous frames adjacent to i-frame in the SSB are required, which could further save most of the decompression time and transmission bandwidth. 

{In short, our SSVC could directly support heterogeneous multimedia analysis tasks just based on partial data decoding, which is achieved and benefited by of semantic-structured coding process and bitstream deployment. We did not jointly train the entire compression framework and subsequent AI application/task models, which is different from previous joint-training based literature~\cite{wu2018compressed,shou2019dmc,zhang2016joint,ma2018joint}}.

Last but not least, we experimentally show how to leverage the semantically structured bitstream (SSB) to better adaptively support downstream intelligent tasks in an adjustable manner (shown in Fig.~\ref{fig:overview} and Fig.~\ref{fig:human_interacion_results}). Such scalable functionality bridges the gap between the video high-efficient compression and machine vision supporting. In summary, the contributions of this paper can be summarized as follows:


\begin{itemize}
    \item We propose an advanced Semantically Structured Video Coding (SSVC) framework to meet the fast-growing requirements of intelligence multimedia analysis. As a new paradigm for intelligent video compression, SSVC could support heterogeneous multimedia analysis tasks just based on partial data decoding, and thus greatly reducing the transmission bandwidth and storage resources. This is achieved by the semantic-structured coding process and bitstream deployment.

    \item In order to efficiently support video downstream tasks based on partially decoded bitstream, we leverage optical flow and residual to describe the dynamic temporal motion information of video, and add them into the semantic-structured bitstream (SSB), which goes beyond the image-based semantic compression framework~\cite{sun2020semantic} and makes our SSVC more general and scalable. We instantiate SSVC framework integrated with action recognition and video object segmentation as a video-based embodiment to reveal the superiority of our method.

    
    
    \item Experimentally, we provide evidences to reveal that our SSVC is more flexible and scalable, which could better adaptively support heterogeneous downstream intelligent tasks with the structured bitstream.


\end{itemize}

The remaining part of this paper is organized as follows: we
introduce recent progress on video compression in Section
II, including traditional hybrid coding pipelines and learning based compression schemes. The details of the proposed Semantically Structured Video Coding (SSVC) framework are introduced in Section III. Comprehensive experiments are conducted and illustrated in Section IV and Section V. We conclude our coding architecture and discuss its future directions in the last section Section VI.

\section{Related Work}

In the current information age, the fast-growing multimedia videos take up most of the daily life of people. It is critical for humans to record, store, and view the image/videos efficiently. For the past decades, lots of academic and industrial efforts have
been devoted to video compression, which aims to achieve a trade-off on the rate-distortion optimization problem. Below we first review the advance of traditional video coding frameworks, as well as the recent booming, developed deep learning based compression schemes. Then, we introduce several task-driven coding schemes on visual data for machine vision in a general sense, revealing its growing importance.

\subsection{Traditional Image/Video Coding Approach}

From the 1970s, the hybrid video coding architecture~\cite{roese1975combined} is proposed to lead the mainstream direction and occupy the major industry proportion during the next few decades. Based on this, the following popular video coding standards have kept evolving through the development of the ITU-T and ISO/IEC standards, including H.261~\cite{xv1989video}, H.263~\cite{sg151996video}, MPEG-1~\cite{iso1993information}, MPEG-4 Visual~\cite{jtccoding}, H.262/MPEG-2 Video~\cite{itu1995generic}, H.264/MPEG-4 Advanced Video
Coding (AVC)~\cite{telecom2003advanced}, and H.265/MPEG-H (Part 2) High Efficiency Video Coding (HEVC)~\cite{sze2014high} standards.

All these standards and improvements based on them\cite{264SP6,enhancingVVC, Mpeg-2SP5,STVVC, WaveletSP7} follow the block-based video coding strategy. Based on this, the intra and inter-prediction techniques are applied based on the corresponding contexts, \ieno, neighboring blocks and reference frames in the intra and inter modes, to remove temporal and spatial statistical redundancies of video frames. However, these kinds of designed
patterns, \egno, block partition, make the prediction only could cover parts of the context information, which
limits its modeling capacity. Besides, the block-wise prediction, along with transform and lossy quantization, causes the blocking effect in the decoding results. As most of the traditional coding architectures generate the bistream in units of the entire image or video, they cannot support partial bitstream decoding or partial objects reconstruction for intelligent video analysis tasks. Besides, different from most of the codecs, the MPEG-4 Visual decomposes video into video object planes (VOPs) and encodes them sequentially. Though MPEG-4 Visual tries to achieve object-oriented bitstream, its implementation must be based on accurate pixel-level segmentation results, which is difficult to achieve at the moment.

\subsection{Learning Based Image/Video Coding Approach}

The great success of deep learning techniques significantly promotes the development of end-to-end learned video coding\cite{Learningbasedcomp, reviewSP1}. For the deep learning based coding methods, they do not rely
on the partition scheme and support full-resolution coding,
which naturally avoids the blocking artifacts. Generally, the representative and powerful feature is extracted via a hierarchical network and jointly optimized with the reconstruction task for high
efficient coding. For instance, the early work~\cite{chen2019learning} focuses on motion predictive coding and proposes the concept of PixelMotionCNN
(PMCNN) to model spatiotemporal coherence to effectively perform predictive coding inside the learning network. Similarly, recurrent neural network~\cite{toderici2017full,toderici2015variable}, VAE generative model~\cite{balle2017end,balle2018variational} and non-local attention~\cite{liu2019practical,liu2019non} are employed to remove the unnecessary spatial redundancy from the latent representations to make features compact, and thus leading to improved coding performance.

For another mainstream branch, lots of efforts are devoted to improving the performance of neural network based video coding frameworks by increasing the prediction ability of
deep networks for intra-~\cite{hu2019progressive} or inter-prediction of video codec~\cite{zhao2019enhanced}. Meanwhile, the end-to-end learned video compression frameworks, such as DVC~\cite{lu2019dvc} and HLVC~\cite{yang2020learning}, further push the compression efficiency up along this
route. All these methods could reduce the overall R-D cost on large-scale video data. Besides, as the entire coding pipeline is optimized in an end-to-end manner, it is also flexible to adapt the rate and distortion to accommodate a variety of end applications, \egno, machine vision analytics tasks.

However, mentioned learning based compression methods typically fail to handle the situation when tremendous volumes of data need to be processed and analyzed fast, because they need to reconstruct the whole picture. The semantics-unknown data still constitutes a major part of the bitstream. So these methods cannot fulfill the
emerging requirement of real-time video content analytics when dealing with large-scale video data. But, these learning based coding frameworks actually provide opportunities
to develop effective VCM architectures to address these challenges.

\begin{figure*}
  \centerline{\includegraphics[width=0.9\linewidth]{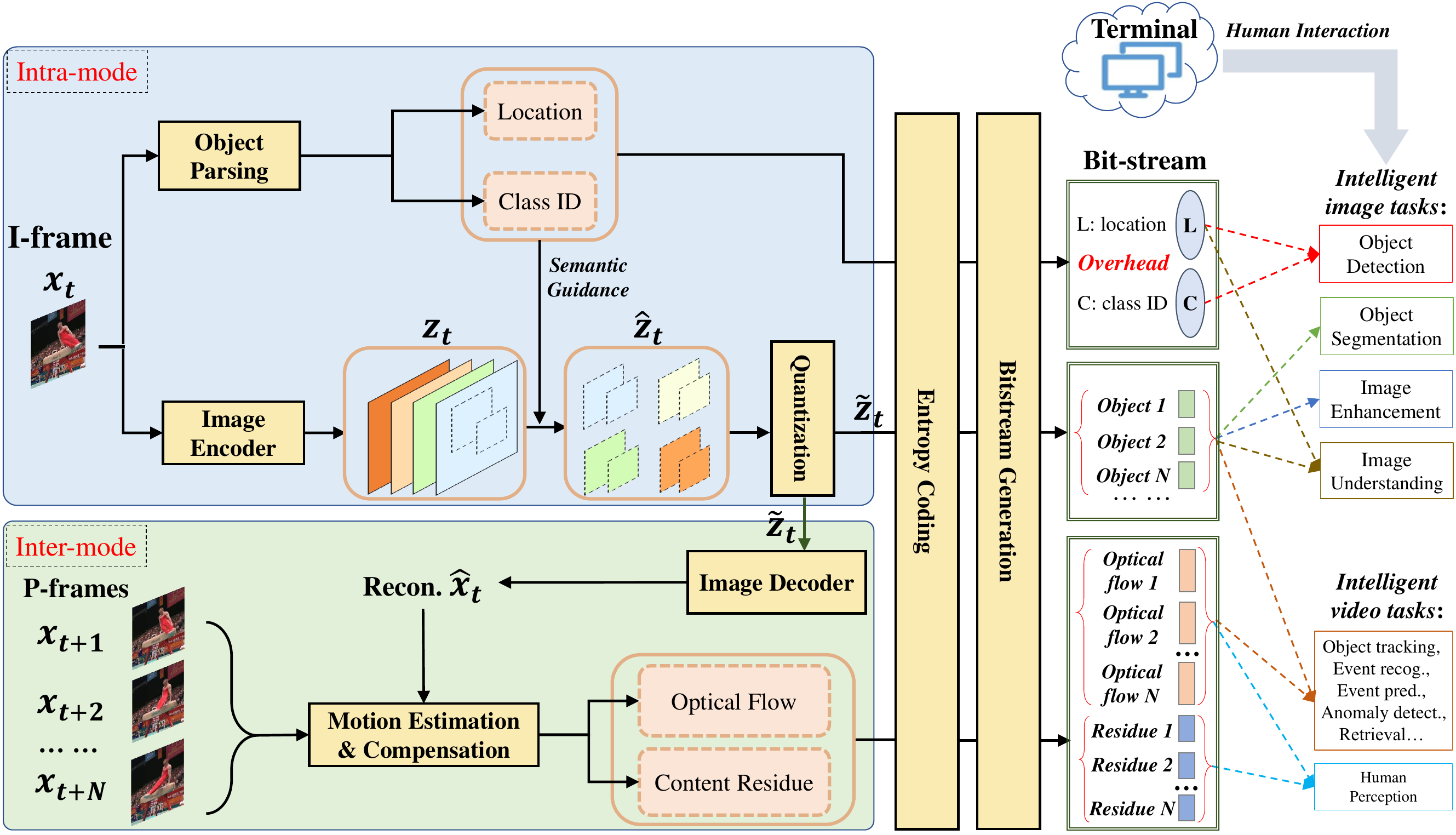}}
    \caption{The overall pipeline of our proposed semantically structured video coding (SSVC) framework, and the illustration of examples of downstream intelligent tasks analytics.}
\label{fig:framework}
\end{figure*}

\subsection{Task-driven Image/Video Coding Approach}

Deep learning algorithms have achieved great success in the actual computer vision tasks, promoting the development of media industry in recent years. Correspondingly, more and more captured videos are directly handled/analyzed by machine algorithms, instead of being perceived by human eyes. Therefore, recent works tend to optimize their compression pipelines according to the feedbacks derived from real task-driven applications rather than the original quality fidelity that aims to meet human perception.

Built upon the traditional codecs, Pu~\etal~\cite{pu2014image} apply
a task-specific metric into JPEG 2000. Liu~\etal~\cite{liu2018deepn} enhance the compression scheme for intelligent applications by minimizing the distortion of frequency features that are important to neural network. CDVS~\cite{pau2013white} and CDVA~\cite{pau2013white} aim at efficiently supporting the search task through compact descriptors using both the traditional method and learning-based method. {Li~\etal~\cite{9472999} implement semantic-aware bit-allocation for the traditional codec based on reinforcement learning.} On the other hand, based on the learning-based coding schemes, Chen~\etal~\cite{chen2019learningFace} propose a learning based facial image compression (LFIC) framework with a novel regionally adaptive pooling (RAP) module that can be automatically optimized according to gradient feedback from an integrated hybrid semantic fidelity metric. These traditional hybrid video coding framework and the aforementioned learning-based methods both
encode the video into binary stream without any semantic
structure, which makes such bitstream failed to directly support intelligent tasks. {Zhang~\etal\cite{zhang2016joint} propose a hybrid content-plus-feature coding scheme framework of jointly compressing the feature descriptors and visual content. A novel rate-accuracy optimization technique is proposed to accurately estimate the retrieval performance degradation in feature coding. {Duan~\etal~\cite{9180095} carry out exploration in the new video coding for machines (VCM) area by building a bridge between feature coding for machine vision and video coding for human vision. They propose a task-specific compression pipeline that jointly trains the feature compression and intelligent tasks.} Ma~\etal~\cite{ma2018joint} provide a systematical overview and analysis on the joint feature and texture representation framework, which aims to smartly and coherently represent the visual information with the front-end intelligence in the scenario of video big data applications.} 


These methods mostly adopt a \textbf{joint training scheme} to not only optimize compression rate but also optimize the accuracy for AI applications. This joint-training based optimization lacks of flexibility, because they need to adjust the compression encoder according to the different subsequent supporting AI tasks. However, in actual applications, it is unrealistic to enforce/adjust the encoder and decoder to be combined with the task. Once their coding framework is well trained on a specific task, it is difficult to adapt it to the other vision tasks.

Therefore, in this paper, we present the concept of semantically structured bitstream (SSB), which contains hierarchical information that represents partial objects existed in the videos and can be directly used for various tasks. Note that, the proposed SSVC video coding framework in this paper is an extension
of our previous image coding pipeline SSIC reported in~\cite{sun2020semantic}. SSVC goes beyond SSIC~\cite{sun2020semantic} on at least four perspectives: 1) SSIC only supports image coding and only could be employed for image-based intelligent anaylstic. On the contrary, our SSVC framework could support image and video coding together, while could be directly employed for image-based and video-based intelligent analysis. 2) the SSB of SSIC only contains static object information of the image, while the counterpart of our SSVC not only encodes static object information contained in the key frames/images, but also integrates motion clues (\ieno, optical flow between neighboring frames) and content residues into bitstream. In general, the SSB of our SSVC framework is compounded with static object semantics information and dynamic motion clues between adjacent video frames. 3) beyond SSIC, we replace the original backbone which is based on a conditional probability model~\cite{mentzer2018conditional} with a stronger VAE-based backbone~\cite{minnen2018joint}, and thus improving the basic compression performance of SSVC. 4) in terms of validation experiments, we add more analysis and experiments on the video-based intelligent tasks, revealing the superiority of SSVC compared to SSIC.

\section{Semantically Structured Video Coding Framework}


In this section, we will introduce the architecture of our proposed Semantically Structured video coding (SSVC) framework. The pipeline is illustrated in Fig~\ref{fig:framework}. In the following sub-sections, we first begin with an overview of the proposed SSVC framework, and then we introduce the details of each component sequentially.


Given a video $X$ that is composed of multiple frames $x_1, x_2 ... x_N$ where $N$ denotes the length of such video clip, the video compression process can be formulated/deemed as a rate-distortion (R-D) optimization (RDO) problem~\cite{sullivan1998rate,kalluri2018adaptive}. The target of such RDO can be understood from two sides, one is minimizing bit-rate cost, \ieno transmission/storage cost, while not increasing fidelity distortion, the other is minimizing distortion with a fixed bit-rate. The Lagrangian
formulation of the minimization RDO problem is given by:
\begin{equation}
    {\rm min}J, \hspace{2mm} where \hspace{1mm} J = R + \lambda D,
    \label{equ:rdo}
\end{equation}
where the Lagrangian rate-distortion functional $J$ is minimized for a particular value of the Lagrange multiplier. More details on Lagrangian optimization are discussed in~\cite{ortega1998rate}. We go beyond the traditional hybrid video coding framework by building up our compression pipeline upon the learning based codecs, in which the modules can be jointly optimized for better implementing R-D optimization. We attempt to define the pipeline of video coding for machine (VCM) to bridge the gap between coding semantic features for machine vision tasks and coding pixel features for human vision.

As shown in Fig.~\ref{fig:framework}, in the compression process, the data encoding has two modes, \emph{intra-mode} and \emph{inter-mode}. Following the traditional hybrid video coding codecs~\cite{hevc} and the exiting learning-based methods~\cite{lu2019dvc,yang2020learning}, we first divide the original video sequence into groups of pictures (GoP). Let $x = \{x_1, x_2,...,x_t, x_{t+1},..., x_N\}$ denote the frames of one GoP unit, where $N$ means the GoP length. Assumed that $x_t$ has been coded by intra-mode, in the next inter-mode coding process, $x_{t+1}, x_{t+2},..., x_N$ is encoded frame-by-frame in a sequential order.

Then, a differentiable quantizer is applied on $\hat{z}_t$ to obtain quantized features $\tilde{z}_t$ to reduce redundant information in the data. After being applied to the entropy coding module, $\tilde{z}_t$ is encoded into the bitstream that can be transmitted or stored. Notably, the extracted high-level semantics (\ieno, location and class information) are also saved into bitstream as overhead, which can be used to directly support downstream intelligent analysis and also guide the partial/specific bitstream decoding (\ieno, partial/specific reconstruction). In summary, the quantized features $\tilde{z}_t$ (can be regarded as low-level content information) and high-level features together constitute the semantically
structured bitstream (SSB).

For the semantically structured bitstream (SSB) deployment, instead of adapting bitstream generation to different downstream intelligent tasks, we pre-define a common/general semantic bitstream deployment. As shown in Fig~\ref{fig:framework}, we divide the bitstream into three groups: 1) header that contains object spatial location and category information, 2) i-frame bitstream that contains different object information, and 3) p-frame bitstream that includes motion clues/information of videos.


\subsection{Intra-mode Coding}
\label{sec:intra}

Intra-mode coding is designed for key frames, \ieno, i-frames of traditional codecs, and can be regarded as a kind of image-based semantics feature compression method. Given a key frame image, that is the $t$-th frame $x_t$ of a video clip $X$, it is first fed into two branches in parallel. One branch employs a feature extractor module to obtain a hidden feature $z_t$, which is semantics-unknown and contains raw content information. The other branch leverages object parsing technique, such as CenterNet~\cite{zhou2019objects}, to extract high-level semantic features from key frame $x_t$, which contains object spatial location information and category information. Such high-level features are not only deployed in the bitstream, but also are used to partition the encoded hidden feature $z_t$ into different groups (\ieno, different spatial areas) $\hat{z}_t$ according to different categories.


\subsubsection{Object Parsing}
\label{intra:obj}

Given $t$-th i-frame, which is noticed as $x_t\in{\mathcal{R}^{W\times{H}\times3}}$ of a video clip $X$. Our goal is to extract semantic features from $x_t$, which are represented by bounding box $(a^{k}_1, b^{k}_1, a^{k}_2, b^{k}_2)$ and class ID $c^{k}$ respectively for object $k$. Following the method in \cite{zhou2019objects}, $x_t$ is first fed into deep layer aggregation (DLA) network~\cite{yu2018deep} to predict a center point heatmap $\hat{Y}\in[0,1]^{{W/R}\times{H/R}\times{C}}$, where $R$ is the output stride and $C$ is the number of predefined object categories. In $\hat{Y}$, a prediction of $1$ corresponds predicted center point of an object, while a prediction of $0$ corresponds to predicted background. Notably, the DLA network can be other fully-convolutional encoder-decoder networks, such as stacked hourglass network~\cite{newell2016stacked,law2018cornernet} and up-convolutional residual networks (ResNet)~\cite{xiao2018simple,he2016deep}. Based on the predicted heatmap, a branch network is introduced to regress the size of all the objects in image $\hat{S}\in\mathcal{R}^{{W/R}\times{H/R}\times2}$. When output stride $R>1$, another additional branch is needed to predict a local offset $\hat{O}\in\mathcal{R}^{W/R\times{H/R}\times2}$ to compensate the error caused by rounding, following~\cite{zhou2019objects}. 

During training stage, the ground truth center point $p\in\mathcal{R}^2$ is converted from bounding box and further mapped to a low-resolution equivalent that is $\tilde{p}=\lfloor{p/R}\rfloor$. Then the ground truth center point is splat to a heatmap version $Y\in[0,1]^{{W/R}\times{H/R}\times{C}}$ using a Gaussian kernel as it does in~\cite{law2018cornernet}. The ground truth of object size is computed as $s_k=(a^{k}_2 - a^{k}_1, b_2^{k} - b_1^{k})$. To optimize the center point heatmap prediction network, we use a penalty-reduced pixel-wise logistic regression with focal loss~\cite{lin2017focal} following~\cite{zhou2019objects}:
\vspace{-0.2cm}
\begin{equation}
    \mathcal{L}_k=-\frac{1}{N}\sum_{abc}
    \begin{cases}
    (1-\hat{Y}_{abc})^\alpha{log(\hat{Y}_{abc})}, \text{if $Y_{abc} = 1$}; \\
        (1-Y_{abc})^\beta(\hat{Y}_{abc})^\alpha log(1-\hat{Y}_{abc}), \text{otherwise},
    \end{cases}
\end{equation}
where $\alpha$ and $\beta$ are hyper-parameters and $N$ is the number of center point in an image. 

The prediction of size and local offset are learned by applying L1 loss respectively:
\begin{equation}
    \mathcal{L}_{size} = \frac{1}{N}\sum^{N}_{k=1}|\hat{S}_{\tilde{p}_k} - s_k|; 
\end{equation}
\vspace{-0.2cm}
\begin{equation}
    \mathcal{L}_{off} = \frac{1}{N}\sum_{p}|\hat{O}_{\tilde{p}} - (\frac{p}{R} - \tilde{p})|.
\end{equation}

Therefore, the total loss function is the weighted sum of all the loss functions with weights $\{1,\lambda_{size},\lambda_{off}\}$.

In inference stage, with the predicted heatmap, the peaks is extracted independently for each category using max pooling operation. With $\hat{P}_c$ denoting the set of $n$ detected center points $\hat{P} = \{(\hat{a}_i, \hat{b}_i)\}_{i=1}^n$ of class ID $c$. Combined with predicted size $\hat{S}_{\hat{a}_i,\hat{b}_i}=(\hat{w}_i,\hat{h}_i)$ and local offset $\hat{O}_{\hat{a}_i,\hat{b}_i}=(\triangle\hat{a}_i,\triangle\hat{b}_i)$, the predicted bounding box can be represented as follow:
\begin{equation}
\begin{aligned}
    (\hat{a}_i + \triangle\hat{a}_i - \hat{w}_i/2,\ \hat{b}_i + \triangle\hat{b}_i - \hat{h}_i/2),\ \\\hat{a}_i + \triangle\hat{a}_i + \hat{w}_i/2,\ \hat{b}_i + \triangle\hat{b}_i + \hat{h}_i/2)).
\end{aligned}
\end{equation}



\subsubsection{Image Compression and Bitstream Disployment}
\label{intra:enc}
\begin{figure*}[t]
  \centerline{\includegraphics[width=0.9\linewidth]{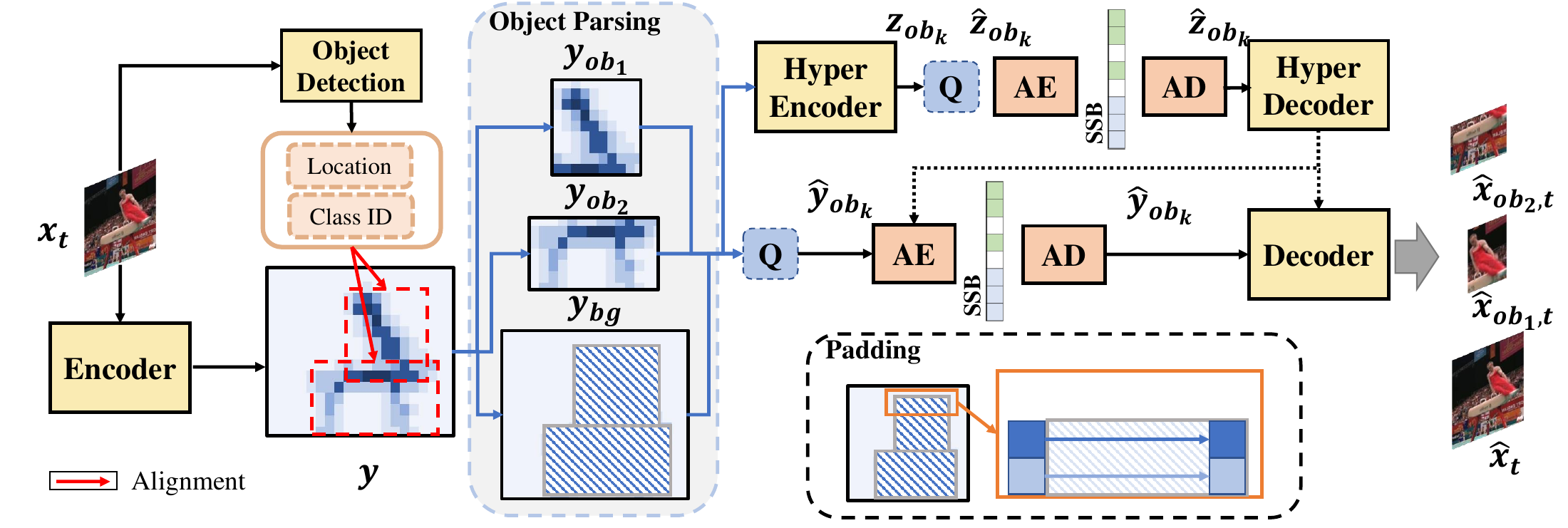}}
  \vspace{-1mm}
    \caption{{Image encoder pipeline.} Encoder, Hyper Encoder, AE and quantization operation are needed in image encoder, while Context Model, Entropy Parameters, AD, Factorized Entropy Model, Decoder and Hyper Decoder are needed in image decoder to recover an image from bitstream.}
\label{fig:img_enc}
\vspace{-2mm}
\end{figure*}

The compression network for i-frame $x_t$ can be divided into two sub-networks as~\cite{minnen2018joint}. One is a core autoencoder (including Encoder and Decoder module), and the other is a sub-network that contains a context model and a hyper-network (including Hyper Encoder and Hyper Decoder module), as is shown in Fig~\ref{fig:img_enc}. 

Specifically, the input $x_t$ is first transformed into latent representation $y$ by Encoder module. Then $y$ is re-organized and quantized as $\{\hat{y}_{ob_1}, \hat{y}_{ob_2},..., \hat{y}_{ob_K}, \hat{y}_{bg}\}$ according to the the $K$ pairs of spatial location information and category information extracted from object parsing branch, in which $\hat{y}_{bg}$ represents the latent representation of background. 
{Then the Arithmetic Encoder (AE) module codes the symbols coming from the quantizer into binary bitstream for each $\hat{y}_{ob_i}$ to generate semantically structured bitstream (SSB), which will be used for storage and transmission. 
Notably, the entropy encoding part of the background $\hat{y}_{bg}$ has been improved in order to minimize the duplication region when encoding the background as~\cite{sun2020semantic}. We fill the inside of the object region with the pixels which are at the left of the border. And in entropy coding, the duplicate parts are coded only once.
The Arithmetic Decoder (AD) could transform bitstream back into the latent representation which can be used for image analysis tasks, and the Decooder also could reconstruct the partial image or the whole image from SSB~\cite{sun2020semantic}.} 

During training stage, only the compression of the whole image is considered, following ~\cite{balle2017end,minnen2018joint}. Then the RDO problem in Equation~\ref{equ:rdo} can further be formulated as following:
\begin{equation}
    \begin{array}{ll}
         R +\lambda\cdot{D}  &=\mathbb{E}_{x\sim{p_x}}[-{\log_{2}{p_{\hat{y}}{(\lfloor{f(x)}\rceil)}}}] \\
         &+ \lambda\cdot \mathbb{E}_{x\sim{p_x}}[d(x,g(\lfloor{f(x)}\rceil))], 
    \end{array}
    \label{equ:RD}
\end{equation}

where $p_x$ is the unknown distribution of natural images, $\lfloor{\cdot}\rceil$ denotes quantization, $f(\cdot)$ and $g(\cdot)$ denote encoder and decoder respectively, $p_{\hat{y}}(\cdot)$ is a discrete entropy model used to estimate the rate by approximating the real marginal distribution of the latent, $d(\cdot)$ is the metric to measure the distortion such as mean squared error (MSE) and MS-SSIM, and $\lambda$ is the Lagrange multiplier to determine the desired trade-off between rate and distortion. 

To estimate the rate for optimization, following ~\cite{balle2017end,minnen2018joint}, the latent $\hat{y}_i$ is modeled as a Gaussian convolved with a unit uniform distribution to ensure a good match between the actual discrete entropy and the continuous entropy model used during training. Then the distribution of latent is modeled by predicting the mean and scale parameters conditioned on the quantized hyperprior $\hat{z}$ and causal context of each latent element $\hat{y}_{<i}$ (\textit{e.g.}, left and upper latent elements). 

The entropy model for hyperprior is a non-parametric, fully factorized density model, as $\hat{z}$ is proved to comprises only a very small percentage of the total bit-rate. 

In inference stage, to generate SSB, given the set of latent from a specific input image $\{\hat{y}_{ob_1}, \hat{y}_{ob_2},..., \hat{y}_{ob_K}, \hat{y}_{bgd}\}$, AE code each of them individually based on their respective hyperprior $\hat{z}_{ob_k}$ (or $\hat{z}_{bg}$) and casual context$\hat{y}_{ob_k,<i}$ (or $\hat{z}_{bg,<i}$). Notably, in order to reduce the coding redundancy caused by re-organization of latent, we introduce two optimization strategies: 1) when objects overlap each other, the union of them is fed into AE; 2) when coding the $\hat{y}_{bg}$, each of the spatially discontinuous part will be padded with the left boundary of the current discontinuous part as~\cite{sun2020semantic}. 




\subsection{Inter-mode Coding}~\label{sec:inter}

The inter-mode coding is designed for continual frames. For inter-mode coding of our SSVC, we focus on low-latency video streaming, which means all inter-frames are coded as p-frame. Given previously decoded frame $\hat{x}_t$ (named as reference frame following traditional codecs), the current frame ${x}_{t+1}$ sequentially perform motion estimation and motion compensation with $\hat{x}_t$ as reference frame. As a consequence, we could get motion clues, \ieno, optical flow and content residue from the encoded frames.

\begin{figure}
  \centerline{\includegraphics[width=0.9\linewidth]{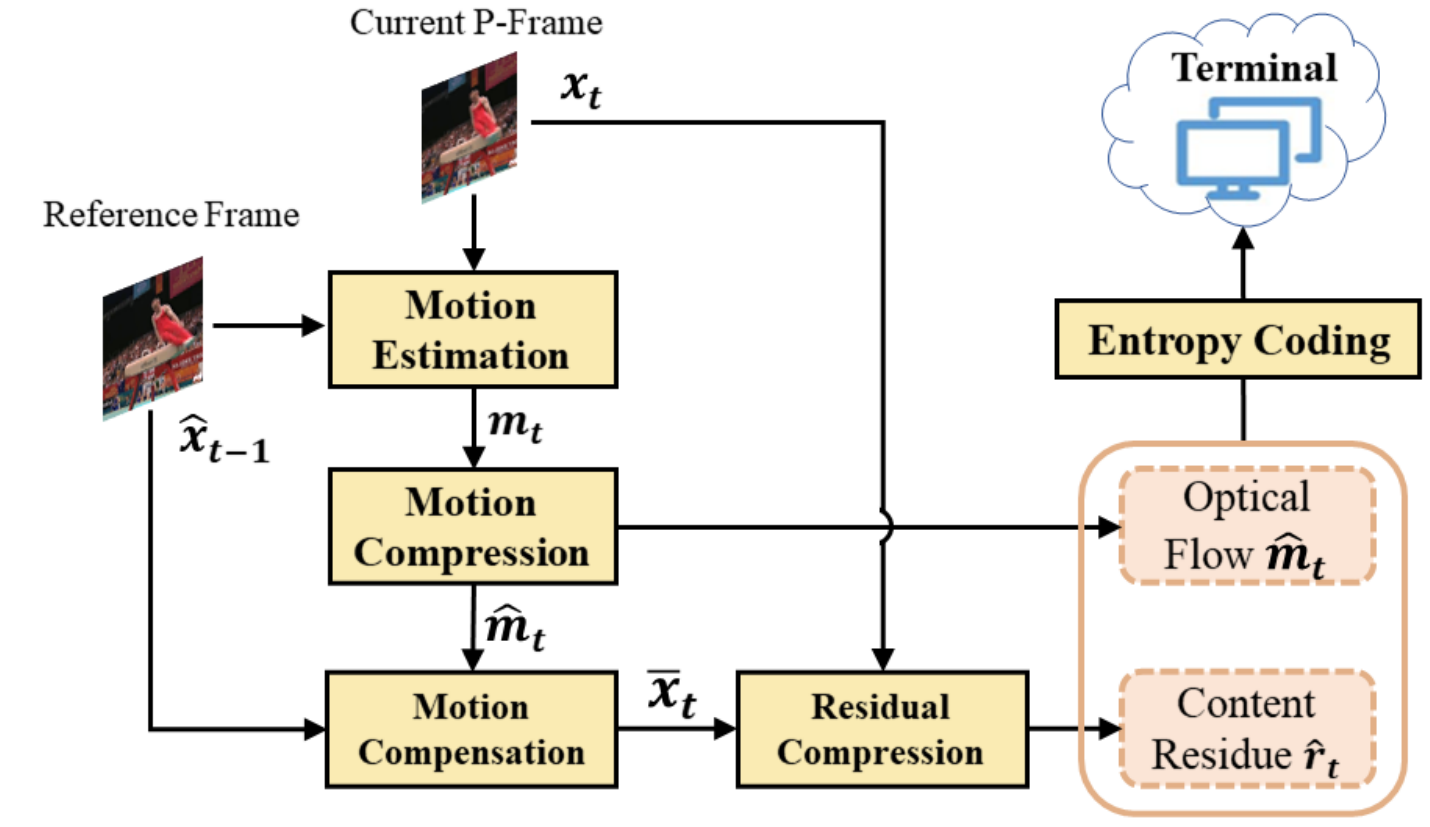}}
  \vspace{-1mm}
    \caption{The overall flowchart of our p-frame compression procedure in the inter coding mode.}
\label{fig:inter_pipeline}
\vspace{-2mm}
\end{figure}

We build our p-frame coding framework based on recent learning-based video coding methods~\cite{feng2020learned}. As shown in Fig.~\ref{fig:inter_pipeline}, the overall coding pipeline contains four basic components: Motion Estimation (ME), Motion Compression (MC), Motion Compensation (MCP) and Residual Compression (RC). We employ the optical flow network PWC-Net~\cite{sun2018pwc} as our ME network. The original output of PWC-Net is in the down-scaled domain with a factor of 4, therefore we upsample it to pixel domain using bilinear interpolation. For the compression of optical flow (\ieno, motion compression), we use the i-frame compression framework and simply change the number of input/output channels. MCP module first warps the reference frame to the current frame by decoded optical flow and then refine the warped frame using a U-Net-like networks.

Given the previously decoded frame $\hat{x}_{t-1}$ and current frame $x_t$, the ME network generates optical flow $m_t$. The MC network, which is similar to our image coding network, first non-linearly maps the optical flow $m_t$ into quantized latent representations and then transforms it back to reconstruction $\hat{m}_t$. The latent representations are encoded into bitstream by entropy coding. After reconstructing $\hat{m}_t$, the reference frame $\hat{x}_{t-1}$ is first bilinearly warped towards the current frame and then refined with a processing network to obtain the motion compensation frame $\overline{x}_t$. Finally, we compress the feature residual between $x_t$ and $\overline{x}_t$ to remove the remaining spatial redundancy, by using the RC network proposed in \cite{feng2020learned}. More details can be seen in \cite{feng2020learned}.

The whole framework is end-to-end trainable. To better adaptively allocate bits between i-frame and p-frame, we optimize the whole model (including our i-frame coding network) for the rate-distortion loss of a GoP:
\begin{equation}
\begin{split}
R + \lambda D
&= \frac{1}{T}\sum_{t=1}^{T}{{R_t} + \lambda \frac{1}{T} \sum_{t=1}^{T}{\mathcal{D}(\boldsymbol{x}_t, \boldsymbol{\hat{x}}_t)}},
\end{split}
\label{eq:RD_VCM}
\end{equation}
where $R_t$ denotes rate,  $\mathcal{D}(\boldsymbol{x}_t,\boldsymbol{\hat{x}}_t)$ denotes distortion and T is the length of the GoP. The rate term for p-frame consists of the rate of optical flow and residual. Note that optical flow and residual are separately encoded into bitstream by two encoder-decoder networks, and therefore can be independently decoded from the corresponding part of bitstream.
In other words, the bitstream of motion information and content information is structured in our coding framework.

\subsubsection{Training Procedure}
It is difficult to train the whole models from scratch using the rate-distortion loss in Eq. (\ref{eq:RD_VCM}). Thus, we separately pretrain the i-frame coding models (intra-mode of SSVC) and p-frame coding models (inter-mode of SSVC). For the pretraining of our p-frame codec, we first fix the weights of the pretrained Motion Estimation (ME) network and then pretrain the Motion Compression (MC) network with the R-D loss of compensation frame $\overline{x}_{t}: R_{t,m} + \lambda_{m} \mathcal{D}(x_t, \overline{x}_{t})$, where $R_{t,m}$ denotes the rate of optical flow, $\mathcal{D}$ is measured using MSE and $\lambda_m$ is empirically set to 512. Later, the weights of the ME network are relaxed and we add the Residual Compression (RC) network for joint training. In the end, we jointly fine-tuned both the i-frame
208 and p-frame models with the proposed R-D loss in Eq. (\ref{eq:RD_VCM}).

\subsubsection{Bitstream Disployment}
As mentioned before and shown in Fig~\ref{fig:inter_pipeline}, the compressed optical flow $\hat{m}_{t}$ and residual $\hat{r}_{t}$ are separately encoded into bit-stream by two encoder-decoder networks, and therefore can be independently decoded from the corresponding part of SSB, which enables our SSVC could directly support more video tasks. For example, only based on objects of key frames (i-frames) and their corresponding motion clues (\ieno, optical flows), terminal users could successfully conduct action recognition. 




\begin{figure*}[t]
  \centerline{\includegraphics[width=1.0\linewidth]{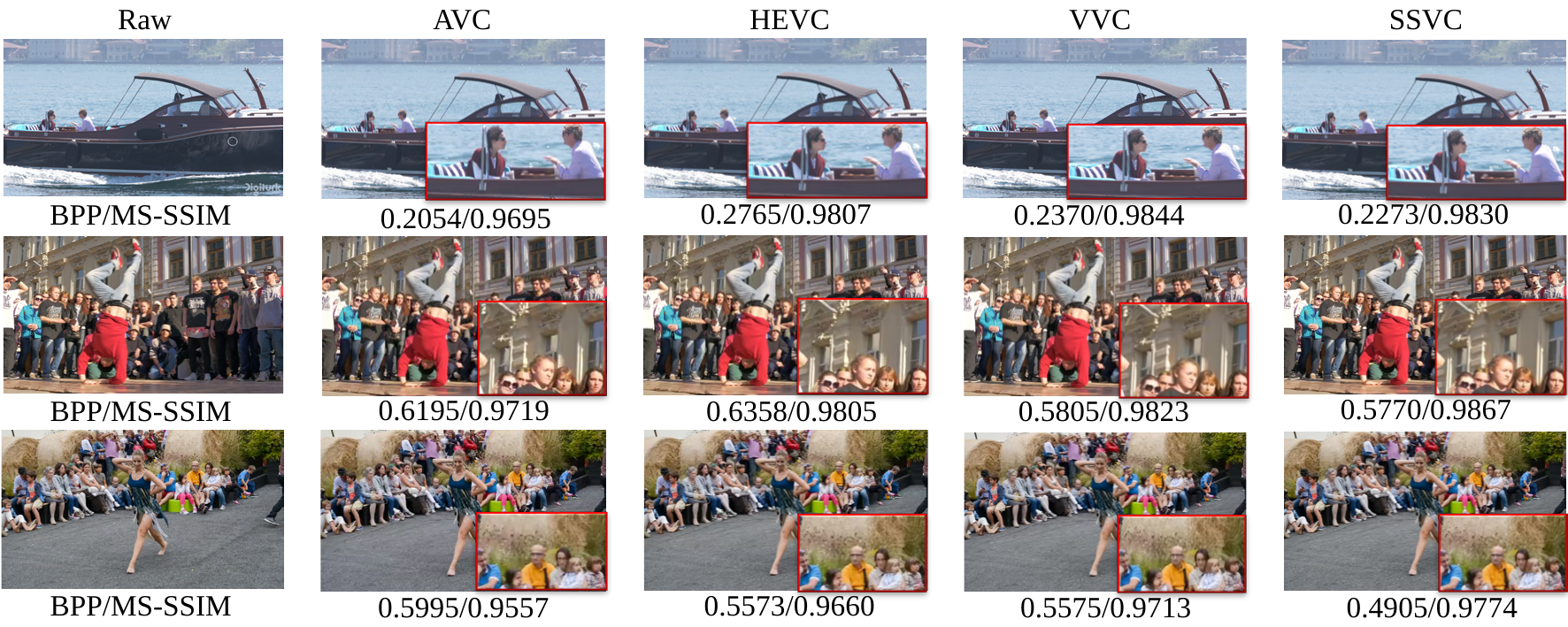}}
  \vspace{-1mm}
    \caption{Qualitative compression performance comparison with the existing compression frameworks.}
\label{fig:qualitative_results}
\vspace{-2mm}
\end{figure*}

\begin{figure*}[htb]
  \centerline{\includegraphics[width=1.0\linewidth]{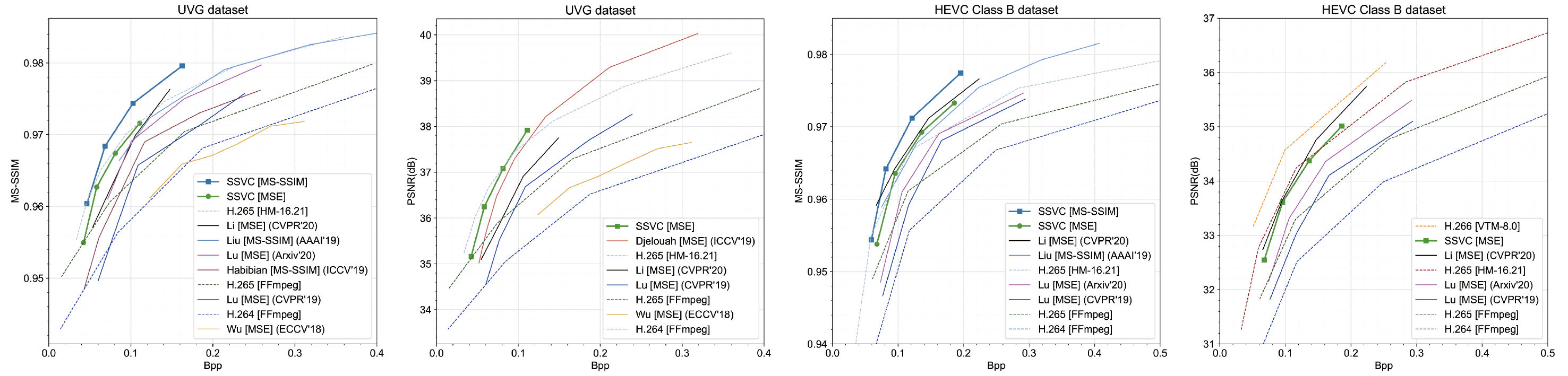}}
  \vspace{-1mm}
    \caption{Quantitative compression performance comparison with the existing learning-based video frameworks and the mainstream traditional video codecs. The curve of SSVC contains all the overheads including class ID and location. }
\label{fig:RD_results}
\vspace{-2mm}
\end{figure*}


\section{Experiments on Compression Performance}

\subsection{Datasets}
To train our i-frame compression models, we use a subset of ImageNet database~\cite{deng2009imagenet}. To train the whole video compression models, we use the Vimeo-90k septuplets dataset~\cite{xue2019video} which consists of 89,800 video clips with diverse content. To report the rate-distortion (R-D) performance~\cite{davisson1972rate,blau2019rethinking}, we evaluate our proposed method on the UVG dataset~\cite{mercat2020uvg} which includes seven 1080p video sequences, and HEVC standard Common Test Sequences~\cite{sullivan2012overview} known as Class B (1920$\times$1080), Class C (832$\times$480), Class D (416$\times$240) and Class E (1280$\times$720). 

\subsection{Evaluation Metrics and Experimental Setup}

We measure the quality of reconstructed frames using both PSNR and MS-SSIM~\cite{wang2003multiscale}. The bits per pixel (bpp) is used to measure the number of coding bits. Following the common evaluation setting in~\cite{lu2019dvc}, the GoP sizes for the UVG dataset and HEVC standard Common Test Sequences are set to 12 and 10, respectively. In most previous methods for learned video compression, the H.264/H.265 is evaluated by using FFmpeg implementation, which performance is much lower than official implementation. In this paper, we evaluate H.265 and H.266 by using the implementation of the standard reference software HM 16.21~\cite{hevc} and VTM 8.0~\cite{vvc}, respectively. We would like to highlight that H.266 [VTM-8.0] is the latest mainstream video coding standard.

\subsection{Implement Details}

We train four models optimized for MSE with different $\lambda$ values (256, 512, 1024, 2048), and four models optimized for MS-SSIM with $\lambda$ values (4, 8, 16, 32). That is, we experimentally get 4 bit-rate points for codec evaluation. The GoP size is set to 6 during the training. We use the Adam optimizer~\cite{kingma2014adam}. In the pretraining procedure, we randomly crop the training data into 128$\times$128 images/video clips and set the learning rate to 5e-5. In the fine-tuning procedure, the crop size is 192$\times$192 and the learning rate is reduced to 1e-5. The batch size is set to 8 and 6 for the two procedures, respectively.

\subsection{Compression Performance Comparison and Analysis}

\subsubsection{Quantitative Analysis}
We evaluate our model with many state-of-the-art video compression approaches, including learning-based coding framework and the traditional video coding methods (\egno, H.264, H.265 and H.266). The compared learning-based video compression approaches include the p-frame based methods of~\cite{lu2019dvc,lu2020content,liu2019learned,lin2020m}, the B-frame based methods of \cite{wu2018video,djelouah2019neural, yang2020learning}, and the transform-based method \cite{habibian2019video}. Among them, \cite{lu2019dvc,lu2020content,wu2018video,djelouah2019neural, lin2020m} are optimized for MSE and \cite{liu2019learned,habibian2019video} are optimized for MS-SSIM.

The corresponding quantitative comparison results are shown in Fig.~\ref{fig:RD_results}, we observe that 1). Our proposed video coding framework significantly outperforms the exiting learning-based video compression methods in both PSNR and MS-SSIM. 2). As an end-to-end learning-based codec, our coding framework outperforms the most mainstream traditional hybrid coding framework--H.266 [VTM-8.0] in terms of MS-SSIM. 3). Compared to H.265 [HM-16.21], our codec that is optimized with MSE provides competitive results in PSNR but achieves better results in MS-SSIM. We analyse that is because the autoencoder-based compression modules are easy to train with MS-SSMI as an objective, and the same trend has been observed in other learning-based codecs~\cite{balle2018variational,minnen2018joint}.


\subsubsection{Qualitative Analysis} The qualitative comparisons with the existing codecs are shown in Fig.~\ref{fig:qualitative_results}, where we deliberately enlarge some regions with complex texture for clear comparison. We can easily observe that the quality of the reconstruction that using our compression framework is much better than that of using AVC (H.264), and comparable with H.265 or H.266.


\section{Experiments on Supporting Intelligent Applications with Partial Bitstream}

\begin{figure}
  \centerline{\includegraphics[width=1.0\linewidth]{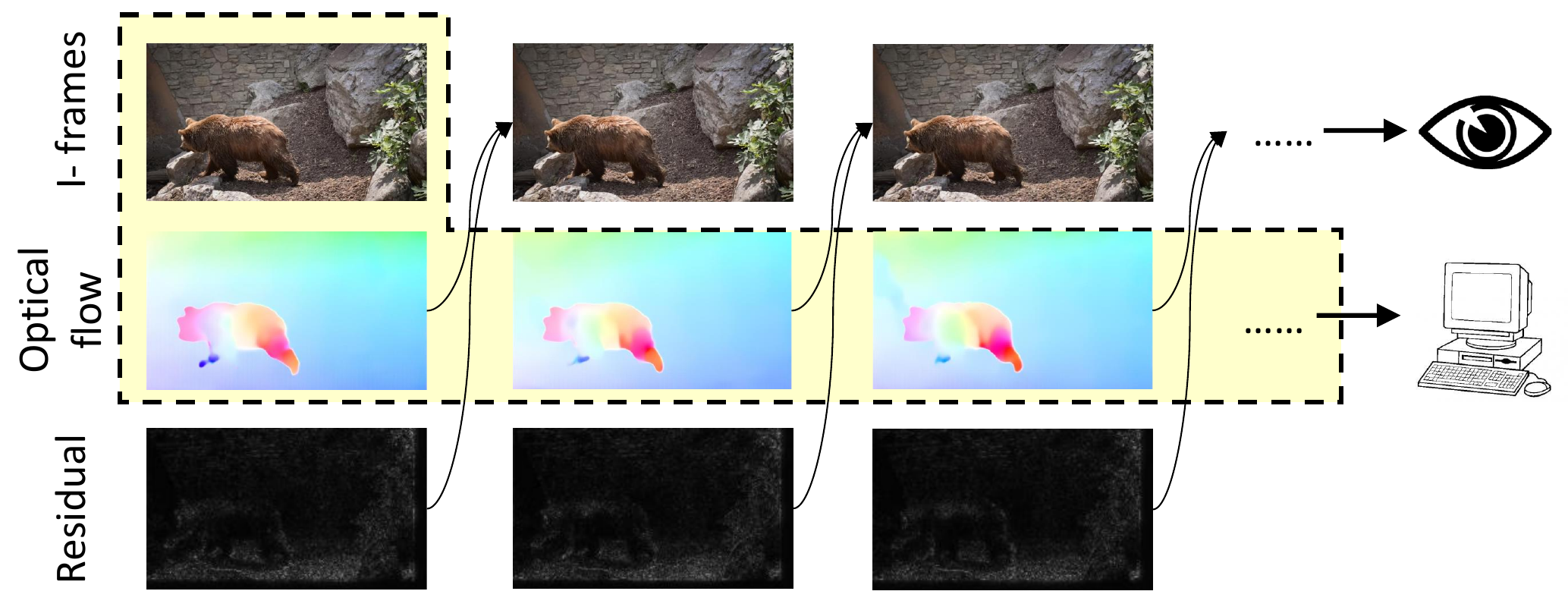}}
  \vspace{-1mm}
    \caption{We use the task of action recognition as an example to illustrate how to adaptively support downstream tasks based on the proposed SSVC framework.}
\label{fig:human_interacion_results}
\vspace{-2mm}
\end{figure}

It is important to show how to leverage the generated semantically structured bitstream (SSB) of our SSVC framework to better adaptively support downstream intelligent machine tasks. Experimentally, in Fig.~\ref{fig:human_interacion_results}, we take the task of video action recognition as an example for explanation. Several i-frames (or partial-decoded i-frames) and several optical flows have already met the needs of most computer vision tasks and can be adjusted in quantity according to the performance-rate trade-off. Also, the whole video can be reconstructed completely for human eyes if needed. Such adjustable decompression could help to achieve an optimal balance between video compression efficiency and intelligent applications supporting.


In summary, considering the practical industrial value and wide application prospects, we take multiple representative heterogeneous machine tasks, including image-based object detection and pose estimation, video-based action recognition and object segmentation, to validate the superiority and scalability of the proposed semantically structured video coding (SSVC) framework and the corresponding semantically structured bitstream (SSB).

\begin{figure}
  \centerline{\includegraphics[width=0.85\linewidth]{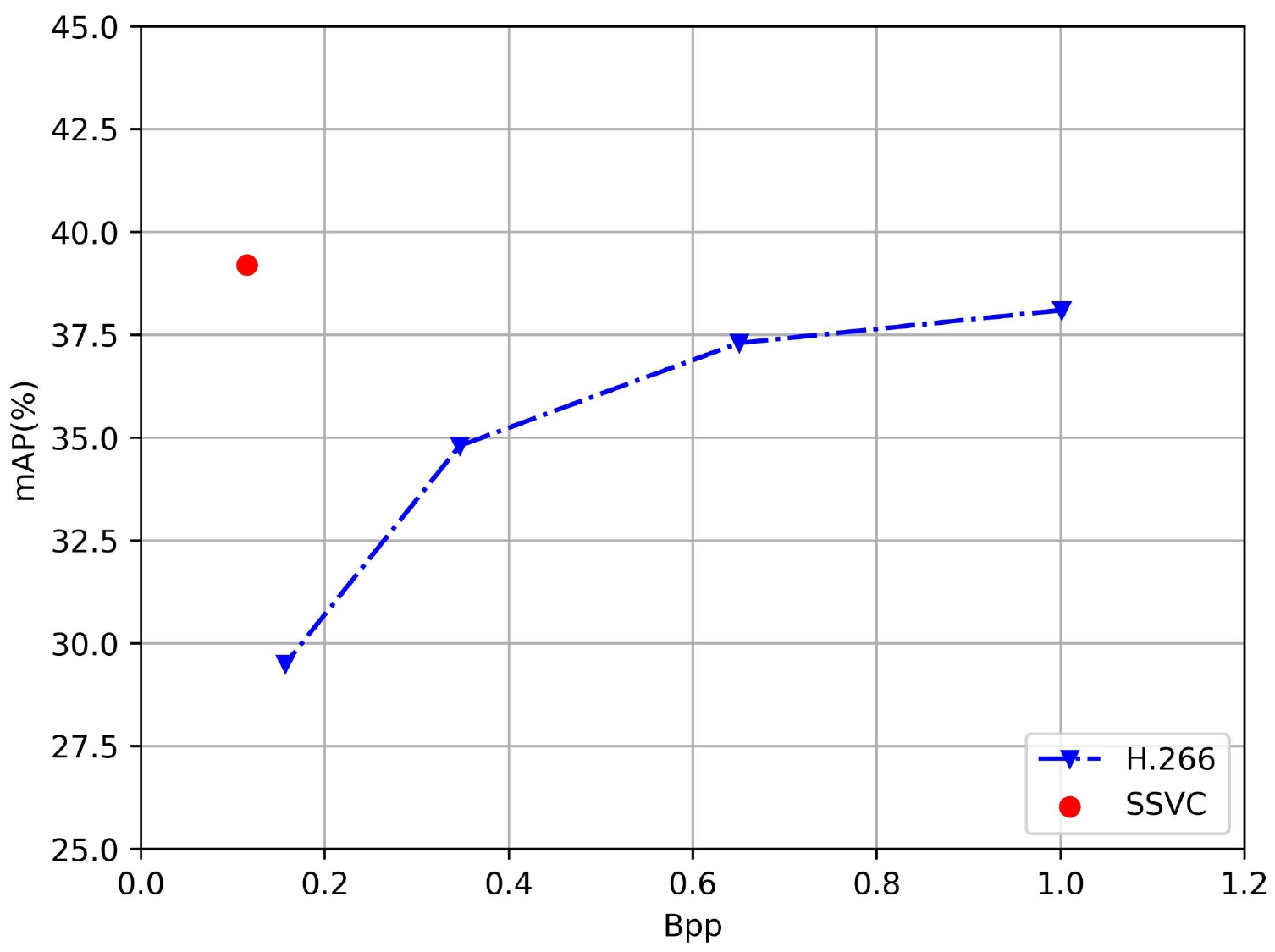}}
  \vspace{-2mm}
    \caption{Performance comparison of compression efficiency \& detection accuracy with traditional video codec H.266. We can see that our SSVC framework do not need to decode all the streams to perform object detection well, which enables our framework to have a higher decoding efficiency and simultaneously gets a satisfactory object detection performance.}
\label{fig:detection}
\vspace{-3mm}
\end{figure}

\subsection{Image-based Downstream Task Evaluation}

\subsubsection{Dataset and Implement Details}
We evaluate our semantically structured video coding (SSVC) framework on object detection. We use COCO2014~\cite{lin2014microsoft} as evaluation dataset. COCO2014 contains 82,783 samples for training and 40,504 samples for validation, in both of which 80 classes/categories are included/covered. Each sample contains at least one object. Here we compare the performance of different coding frameworks on minival set, which is a subset selected from the validation set, containing 5,000 samples.

The intra codec of SSVC is realized with PyTorch. The model is trained on ImageNet dataset with Adam optimizer and a learning rate of 1e-5 for 150w iterations. During the training, we randomly crop the input image into a patch of $256\times256$ and the batch size is 8. When testing, the input needs to be padded into an image whose length and width are both multiples of 64.


\subsubsection{Results of Detection}
In this subsection, we evaluate the performance of our SSVC codec on the object detection task. We compare with the mainstream traditional codec -- H.266 intra codec. For evaluating both of our framework and H.266, the officially released object detection network CenterNet~\cite{duan2019centernet} {with HG network} is adopted as a critic. For VVC codec, we test the performance with QP set as 27, 32, 37 and 42. Correspondingly, We test the performance of our SSVC with 4 different bit-rates, where the $\lambda$ is set as 192, 512, 786 and 1024. The result is shown in Fig.~\ref{fig:detection}, the performance of detection task is evaluated by mAP with intersection of union (IoU) set as $0.50:0.95$. We have the following observations: thanks to the object detection network/function that included in the encoder of our SSVC intra-mode codec, the object detection results (\ieno, bounding box positions) are coded as a part of the semantically structured bitstream (SSB). Therefore, our SSVC framework could directly perform object detection without decoding all the bitstream, which enables our framework to have a higher decoding efficiency and promises a better object detection performance. {Note that, the execution of the object detection task with our semantically structured bit-stream is a special case. For the detection task we can decompress the detection result directly from the header and this part is independent of compression rate. That is because that the IDs and bboxes happend to be compressed without loss and stored in the header}, which are extracted in the encoder side for decomposition and recombination of the latent features. Therefore, the dot in the Fig.~\ref{fig:detection} represent the BPP of the header and the accuracy of the IDs and bboxes that is already present in the header.


\begin{figure}
  \centerline{\includegraphics[width=0.9\linewidth]{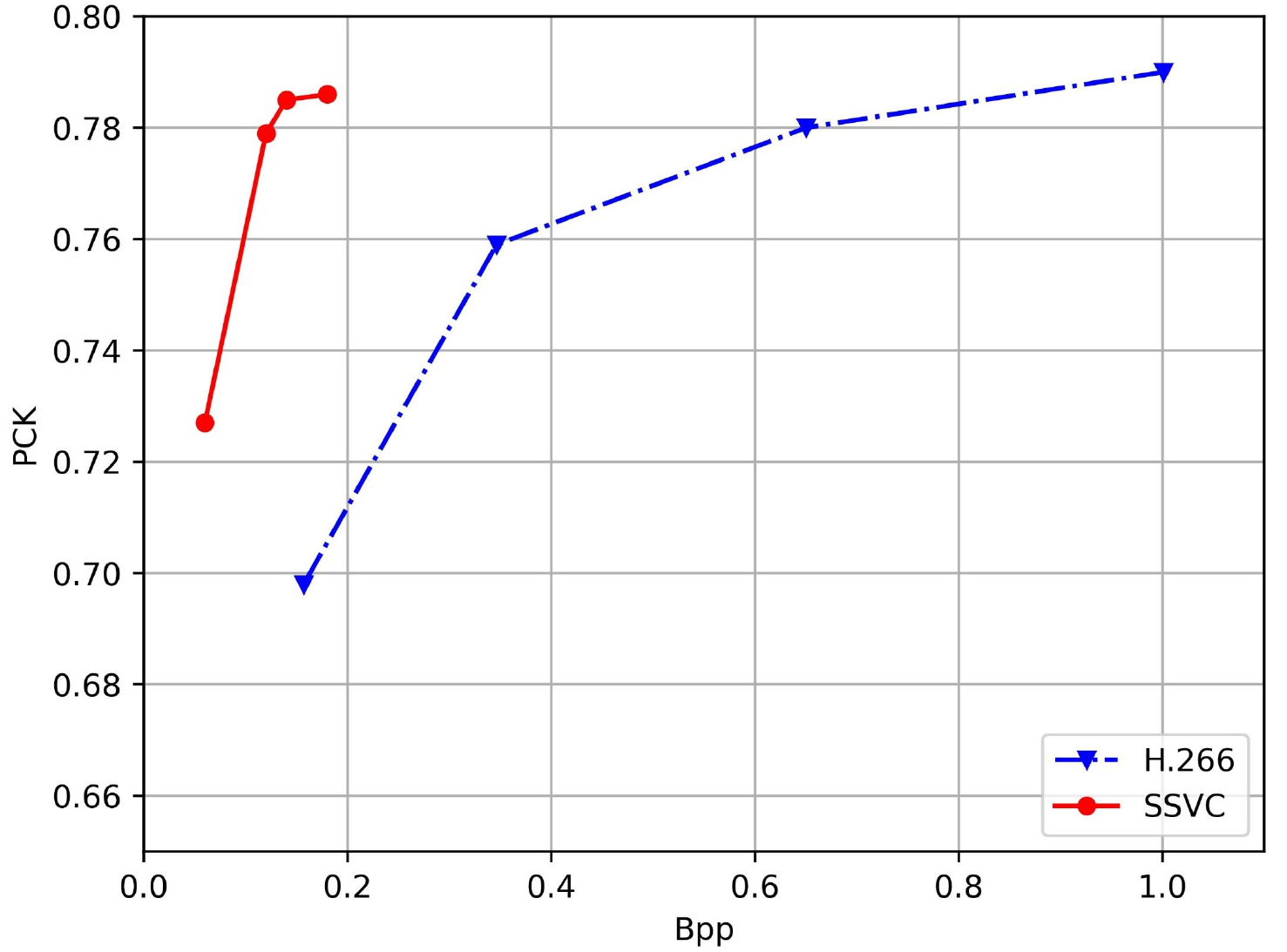}}
  \vspace{-1mm}
    \caption{Performance comparison of compression efficiency \& pose estimation accuracy with traditional video codec H.266. We can see that our SSVC framework achieves a higher decoding efficiency with partial decompression, and promises a better pose estimation performance at the same time.}
\label{fig:pose}
\vspace{-2mm}
\end{figure}

\subsubsection{Results of Pose Estimation}
Based on COCO2014 dataset, we further take the pose estimation task to indicate the superiority of the proposed SSVC coding framework. Specifically, SSVC supports the partial bitstream decompression, and we take these decompressed partial images to conduct pose estimation. Similarly, we also take the VVC (H.266) codec as an anchor to compare. The pose estimation network is the stacked hourglass network~\cite{newell2016stacked}. The QP setting is consistent with that of object detection task. During the training, we omit all the data augmentation techniques and just train models from scratch on the original COCO2014 dataset. We use the RMSprop optimizer with learning rate set as 0.0025. All the decompressed images are resized to $256\times256$. 

{As described in the methodology section of \emph{Intra-mode Coding}, the high-level information (ID and bbox) are stored in header and corresponding low-level feature is stored as well in our semantic structure bitstream (SSB), we can search the entire bitstream to find bitstream related to person. The partial bitstream is first entropy-decoded to latent feature and fed into decoder to obtain pixel-level reconstruction. Then the partially decompressed images in which almost only person is included can be fed into pose estimation task.} In the inference stage, we take the PCK (percentage of correct keypoints) as a metric to evaluate the performance of different schemes. Results are shown in Fig.~\ref{fig:pose}, we can observe that the proposed SSVC framework greatly improves the coding efficiency. That's because our SSVC framework could support partially decoding out these task-specific regions (\ieno, the regions contain human body skeleton), which saves a large transmission bit cost in comparison with the fully-decompressed VVC.


\subsection{Video-based Downstream Task Evaluation}

Except for image-based downstream tasks, our SSVC also could directly support heterogeneous downstream video-based intelligent tasks with the dynamic motion information included SSB. To prove that, we use two classic/representative video tasks of video action recognition and video object segmentation to evaluate SSVC.

\subsubsection{Dataset and Implementation Details} 

We evaluate our semantically structured video coding (SSVC) framework for video-based action recognition on a widely-used dataset UCF-101~\cite{soomro2012ucf101}, which contains 13,320 video clips (mostly shorter than 10 seconds) that cover 101 action categories. Each video clip is annotated with one exact action label.

Due to the original videos of UCF-101 are all in AVI format, we first utilize FFmpeg tool to extract frames (\ieno, RGB images) from raw videos. And then, we leverage PWC-Net~\cite{sun2018pwc} to generate the corresponding optical flow for each frame. Following the previous temporal segment action recognition network TSN~\cite{wang2018temporal}, we train two independent CNNs for RGB image and optical flow, respectively. The backbone of both streams is ResNet-152~\cite{he2016deep}. The entire model, \ieno, TSN, are first pretrained on the ImageNet dataset, then finetuned on the UCF-101 using Adam~\cite{kingma2014adam} optimizer with a batch size of 64. The learning rate starts from 0.001 and drops by 0.1 when the accuracy has stopped rising and such a trend has been kept for several training epochs. We leverage color jittering and random cropping for data augmentation. In the inference phase, we take the average accuracy score of five tests as the final action recognition results.

{When comparing to the traditional codecs where optical flows are used for supporting some AI applications, all frames need to be reconstructed at first and then the optical flow can be estimated. For fairness, we use the same tool (PWC-Net) to estimate optical flow for both traditional codecs and SSVC, all the predicted results are inferenced by the same model trained on uncompressed optical flow.}

For the task of video object segmentation, we evaluate our proposed SSVC video coding framework on the DAVIS-16~\cite{perazzi2016benchmark} dataset, which contains 50 high-resolution videos with 3,455 frames in total, where 30 sequences for training and 20 sequences for online validation. The task of video object segmentation requires segmenting all the object instances from background for each video sequence. Note that, the segmented result/mask for the first frame of each video sequence has been provided in the setting of this task.

We use the OSVOS network \cite{caelles2017one} as segmentation backbone, which is first pretrained on the ImageNet~\cite{5206848}, and then trained on the DAVIS training set. In the end, for each test sequence, OSVOS would be fine-tuned on the provided segmented results that correspond to the first frame. 

During the evaluation, all the video frames that are reconstructed through the traditional codecs are directly sent into the OSVOS network. For the proposed SSVC framework, in each GoP, we only need to decode out the key frames (\ieno, i-frames) of video as input to generate the corresponding binary masks, then we refine these masks according to the decoded optical flows of p-frames through a simple mask refinement module based on U-Net~\cite{ronneberger2015u}, which is inspired by~\cite{perazzi2017learning} and please refer to more details from that. Note that, for the video object segmentation task, we replace the PWC-Net with the SOTA RAFT~\cite{teed2020raft} for more accurate optical flow estimation, which is important for this pixel-level task. All experimental settings are consistent for a fair comparison.

\begin{figure}
  \centerline{\includegraphics[width=0.9\linewidth]{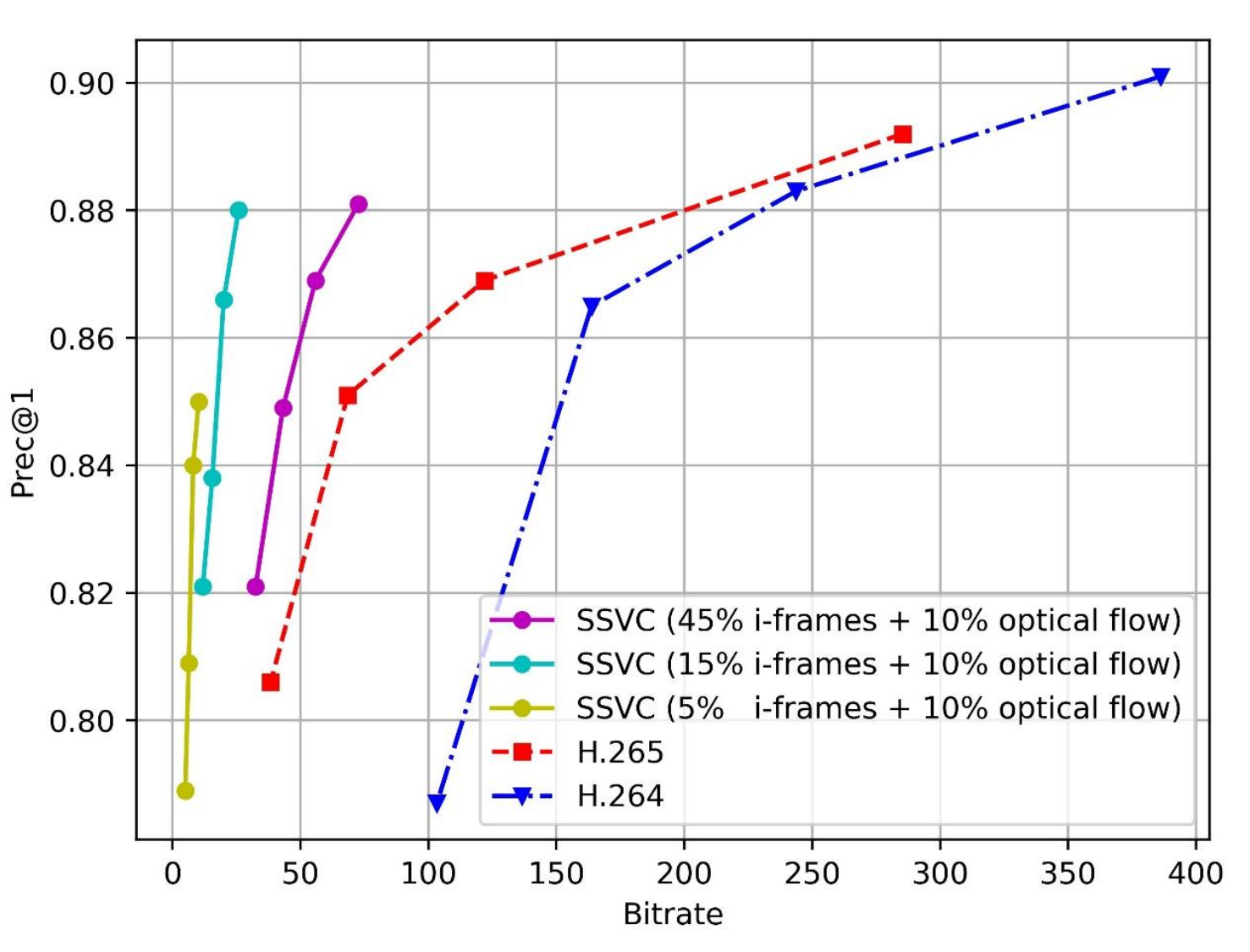}}
  \vspace{-1mm}
    \caption{Performance comparison of compression efficiency \& action recognition accuracy with two traditional video codecs H.264 and H.265. We can see that our coding framework achieves a better compression efficiency and a better recognition accuracy at the same time with partial bitstream decoding. Besides, the whole decoding computational cost of full video reconstruction is also reduced.}
\label{fig:action_recognition}
\vspace{-2mm}
\end{figure}

\subsubsection{Results of Action Recognition}

The proposed SSVC coding framework could directly provide both i-frames and optical flows of p-frames without fully decompression. Thus, when evaluating the compression performance of the proposed SSVC codec on the action recognition task, the RGB stream of TSN takes the partially decoded i-frames (\egno, 5\%, 15\%, and 45\% i-frames) as input and the optical flow stream of TSN take the partial decoded optical flow (\egno, 10\% optical flow) as input. We will prove that such design could achieve a better trade-off between the decompression computational cost and the action recognition accuracy in the following sections.

For the traditional hybrid video coding framework, it is infeasible to only reconstruct the partial i-frames and optical flow, because their encoded bit-streams are semantic-unknown. To get a satisfactory action recognition result, we have to decode the whole video sequences at first, and then estimate the optical flow using PWC-Net, which process is both time-consuming and bandwidth-wasting.

To indicate the superiority of our framework in terms of compression efficiency and recognition accuracy, we employ two popular traditional video codecs, \ieno, H.264 and H.265, as competitors for comparison. We test the performance with several QP settings to make the comparison curve easy/clear to understand/read.

The performance comparison of compression efficiency \& action recognition accuracy is shown in Fig.~\ref{fig:action_recognition}, from which we can observe that, 1). Compared to fully decompressed video codecs H.264 and H.265, directly performing action recognition on the partial bitstream (\egno, 5\%/15\%/45\% i-frames and 10\% optical flow) based on the proposed SSVC framework achieves a better compression efficiency and a better action recognition accuracy at the same time. 2) As the bit-rate increases, the action recognition performance of H.264 and H.265 is gradually approaching. We analyse that is because the raw videos of UCF-101 dataset are lossy. Thus, with the bit-rate increasing, the quality of reconstructed videos would not be further improved. 3) For the proposed SSVC coding framework, the action recognition performance of \emph{SSVC (15\% i-frames + 10\% optical flow)} and \emph{SSVC (45\% i-frames + 10\% optical flow)} is very similar. We analyse that with the content information derived from i-frames increasing, the action recognition task will not be consistently influenced/affected since the motion clues (\ieno, optical flow) is more important when increasing the high bit-rate.

\begin{table}
\centering
\caption{The compression performance (BD-rate, \%) of different schemes. Note that we use the average bit-rate that calculated over the entire action recognition dataset (\ieno, UCF-101) to get this BD-rate saving.}
    \begin{tabular}{c|cc}
    \hline
                        & vs. H.264 & vs. H.265 \\ \hline
    SSVC (45\% i-frame + 10\% flow)    & -69.31     & -40.41     \\
    SSVC (15\% i-frame + 10\% flow) & -88.12     & -76.75     \\
    SSVC (5\% i-frame + 10\% flow)    & -94.06    & -84.25    \\ \hline
    \end{tabular}
    \label{tab:BD-rate-AR}
    \vspace{-2mm}
\end{table}

Besides, since the rate-distortion performance is the key performance indicator for video coding, the widely accepted BD-rate metric~\cite{bjontegaard2001calculation} is also adopted in our experiment. Note that here the ``distortion'' metric is replaced with ``recognition accuracy'', which measures the equivalent bit-rate change (negative means performance improvement and the lower the better) under the same recognition accuracy. 


Table~\ref{tab:BD-rate-AR} demonstrates the BD-rate results on the entire action recognition video dataset (\ieno, UCF-101): compared with the traditional codecs H.264 or H.265, the proposed SSVC could achieve on average over 40\% BD-rate saving. This is a quite significant improvement in video coding research area, since it usually can reach 50\% BD-rate saving every 10 years~\cite{sullivan2012overview} under the traditional hybrid coding framework.

\subsubsection{Results of Video Object segmentation}


Thanks to the SSVC framework could separately decode out both i-frames and optical flows of p-frames without fully decompression, we set up multiple cases to comprehensively evaluate the compression \& segmentation performance of our SSVC framework. \emph{SSVC (One i-frame each GoP)}: only use one i-frame and all the optical flow of p-frames in each GoP to perform segmentation. \emph{SSVC (One i-frame every two GoPs)}: only use one i-frame in each two GoPs and all the optical flow of p-frames to perform segmentation. \emph{SSVC (Only optical flow)}: only use all the optical flow of p-frames to perform segmentation. Note that we could conduct object segmentation only using optical flow since the segmented result/mask for the first frame of each video sequence has been provided.

\begin{figure}
  \centerline{\includegraphics[width=0.9\linewidth]{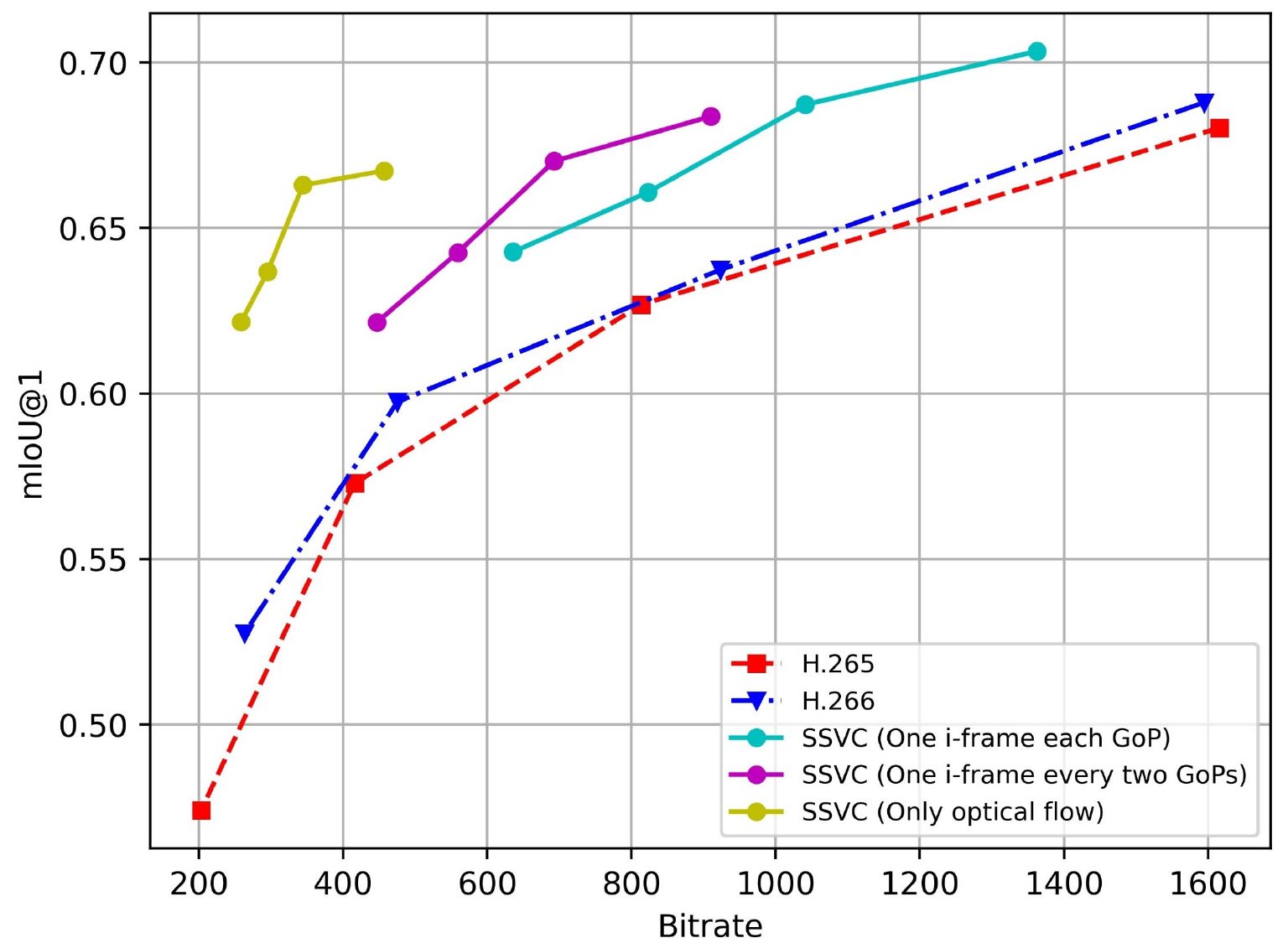}}
  \vspace{-1mm}
    \caption{Performance comparison of compression efficiency \& video object segmentation mIoU with two traditional video codecs H.265 and H.266. Our SSVC performs worse than the other two when reconstructing all video frames but much better with the setting of transferring only partial i-frames and all optical flows.}
\label{fig:video_object_segmentation}
\vspace{-2mm}
\end{figure}

The results are shown in Fig.~\ref{fig:video_object_segmentation}, we have the following observations: 1) When directly using the complete reconstructed video frames to perform object segmentation, H.266 is slightly better than H.265. 2) When only decoding out partial i-frames and all optical flows (\ieno, the bottom three SSVC variants shown in Fig.~\ref{fig:video_object_segmentation}), the proposed SSVC framework performs much better than traditional codecs, achieving a superior trade-off between compression efficiency and segmentation accuracy. 3) When only decoding out the optical flows for segmentation that just cost a little bitstream (\ieno, the scheme of \emph{SSVC (Only optical flow)}), the bit-rate is pretty low but still achieve a satisfying segmentation performance. 

Moreover, the coding backbone of SSVC remains a large improvement space since the learning-based video coding technique is going through a fast development. Thus, we believe that the global performance of SSVC w.r.t the video object segmentation task could be further improved from at least two aspects: 1). using the more advanced video coding backbone. 2). the estimation of optical flows in SSVC is optimized with the R-D objective constrain, which may not be consistent with the ``true motion'' of video objects~\cite{chen2006fast}, leading inaccurate segmentation results. 


Table~\ref{tab:BD-rate-OS} demonstrates the BD-rate saving for video object segmentation task. We observe that, compared with the traditional codecs H.265 or H.266, the proposed SSVC variants all consistently achieve obvious BD-rate saving.

\begin{table}
\centering
\caption{The compression performance (BD-rate, \%) of SSVC with different schemes for the video object segmentation task.}
    \setlength{\tabcolsep}{0.6mm}{
    \begin{tabular}{c|cc}
    \hline
             Description      & vs. H.265 & vs. H.266 \\ \hline
    One i-frame each GoP    & -35.96     & -34.37     \\
    One i-frame each two GoPs & -56.39     & -54.28     \\
    Only optical flow    & -69.81    & -69.29    \\ \hline
    \end{tabular}}
    \hspace{-6mm}
    \label{tab:BD-rate-OS}
\end{table}




\section{Conclusion}

As a response to the emerging MPEG standardization efforts
VCM, in this paper, we propose a learning-based semantically structured video coding (SSVC) framework, which formulates a new paradigm of video coding for human and machine visions. SSVC encodes video into a semantically structured bitstream (SSB), which includes both of the static object semantics characteristics and dynamic object motion clues. The proposed SSVC coding framework with a well-designed SSB has the capability of explicitly supporting the heterogeneous intelligence multimedia analytics without fully decompression. Extensive experiments on multiple benchmarks demonstrate that the proposed SSVC framework not only has a comparable basic compression performance compared to mainstream video coding schemes, but also could directly support intelligent tasks with a large computational cost saving.

\section*{References}

\bibliography{main}

\end{document}